# Digital Twins for Patient Care via Knowledge Graphs and Closed-Form Continuous-Time Liquid Neural Networks

Logan Nye, MD[1,2]
[1]SORG Orthopaedic Research Group
[2]Department of Orthopaedic Surgery
Massachusetts General Hospital

*Abstract*— **Digital twin technology has is anticipated to transform healthcare, enabling personalized medicines and support, earlier diagnoses, simulated treatment outcomes, and optimized surgical plans. Digital twins are readily gaining traction in industries like manufacturing, supply chain logistics, and civil infrastructure. Not in patient care, however. The challenge of modeling complex diseases with multimodal patient data and the computational complexities of analyzing it have stifled digital twin adoption in the biomedical vertical. Yet, these major obstacles can potentially be handled by approaching these models in a different way. This paper proposes a novel framework for addressing the barriers to clinical twin-modeling created by computational costs and modeling complexities. We propose structuring patient health data as a knowledge graph and using closed-form continuous-time "liquid" neural networks (CfCs) for real-time analytics. By synthesizing multimodal patient data and leveraging the flexibility and efficiency of CfCs and knowledge graph ontologies, our approach enables real-time insights, personalized medicine, early diagnosis and intervention, and optimal surgical planning. This novel approach provides a comprehensive and adaptable view of patient health along with real-time analytics, paving the way for digital twin simulations and other anticipated benefits in healthcare.**

*Keywords—informatics, artificial intelligence, deep learning, decision support, clinical model, precision medicine,*

## I. INTRODUCTION

Digital twins are an emerging technology poised to revolutionize every sector of industry.

These same principles can logically carry to healthcare as well. Digital twins are expected to enable precision medicine, personalized therapeutics, earlier diagnoses, and optimized treatment outcomes via simulation capabilites [1]. So far, however, they are still largely absent. While other verticals like manufacturing, systems engineering, infrastructure, and others have begun realizing these advantages, healthcare is more challenging. The computational demands and inherent challenges of modeling complex pathologies leads to steep expenses and missed opportunities. We propose a novel method of modeling a digital twin for patient care that could finally enable this idea. Our approach combines two powerful technologies: (1) knowledge graph representations [2], and (2) closed-form continuous-time liquid neural networks (CfCs) [3]. Together, they provide a comprehensive and adaptable view of individual patient health data that can also be efficiently reflected in real-time, ultimately improving bringing us closer to our end-goal: a true digital twin for patient care.

## II. DIGITAL TWIN TECHNOLOGY

### A. What Are Digital Twins?

Digital twins are virtual representations of physical entities, processes, or systems that leverage real-time data and advanced analytics to optimize performance, enable predictive maintenance, and drive innovation. In the healthcare industry, digital twins can revolutionize patient care, clinical research, and healthcare infrastructure management. The concept of digital twins first arose at NASA and gained traction after Gartner named digital twins as one of its top 10 strategic technology trends for 2017 [1]. They estimated that by 2020, 21 billion connected sensors and endpoints would enable digital twins for billions of things [2].

### B. Digital Twins in Healthcare

A digital twin's architecture comprises three core components: the physical entity, the virtual representation, and the data management system [3]. These components are interconnected through an IoT (Internet of Things) network, enabling real-time synchronization and communication. In healthcare, the physical entity can range from an individual patient to an entire hospital. Data is collected through sensors, wearables, or other IoT devices, capturing vital parameters such as heart rate, blood pressure, and body temperature [4]. The virtual representation is a data-driven model that mirrors the physical entity's behavior, state, and properties. It is created using a combination of mathematical models, machine learning algorithms, and simulation techniques [5].

## C. Twins-enabled Healthcare Analytics

Mathematical models are utilized to represent complex biological systems, such as organ function, metabolic pathways, or disease progression [6]. These models can be deterministic (e.g., differential equations) or stochastic (e.g., Monte Carlo simulations). Advanced analytics techniques, such as predictive analytics, prescriptive analytics, and real-time analytics, are applied to derive insights and facilitate decision-making [7]. These techniques often involve the application of machine learning and artificial intelligence algorithms, including supervised, unsupervised, and reinforcement learning techniques, which are employed to analyze and predict patterns within the collected data [8]. Deep learning architectures, such as convolutional neural networks (CNNs) and recurrent neural networks (RNNs), have shown promising results in modeling patient-specific medical conditions [9]. In addition, simulation techniques, like agent-based modeling and discrete event simulation, can be used to recreate the dynamic behavior of healthcare processes or systems, allowing for the evaluation of various intervention strategies and resource allocation scenarios [10,11].

## D. A Valuable Clinical Tool

Healthcare digital twins require secure, scalable, and efficient storage solutions to handle vast amounts of structured and unstructured data [12]. Technologies like distributed file systems (e.g., Hadoop Distributed File System) and cloud-based storage services are commonly used. Data processing involves cleaning, transforming, and integrating data from multiple sources to ensure accuracy and consistency. Techniques like data wrangling, normalization, and feature extraction are employed in this stage [13]. Digital twins are valuable to healthcare because they facilitate personalized medicine [14], enable early diagnosis and intervention [15], enhance clinical trials [16], support remote patient monitoring [17], optimize healthcare facilities [18], enable predictive maintenance [19], and improve medical education and training [20]. These benefits contribute to improved patient outcomes, operational efficiency, and cost savings in the healthcare industry.

## III. KNOWLEDGE GRAPHS

### A. What are Knowledge Graphs?

Knowledge graphs are a graph-based data structure for representing structured and semi-structured data [4]. By encoding entities as nodes and relationships as edges, knowledge graphs allow for complex querying and reasoning over large datasets, making them a natural choice for representing patient health data [5]. Medical ontologies, such as the Systematized Nomenclature of Medicine - Clinical Terms (SNOMED CT) [6] and the Human Phenotype Ontology (HPO) [7], provide the foundation for our knowledge graph representation, ensuring accurate and consistent terminology.

### B. Problems Solved by Knowledge Graphs

Knowledge graph is a modern and effective method for harmonizing enterprise data, allowing representation and operationalization

Knowledge graph is a modern and effective method for harmonizing enterprise data, allowing representation and operationalization of data, data sources, and databases of all types. As we progress further into the AI era, the increasing amount of data is being utilized for business benefits and advantages, steadily transforming data into knowledge [32]. In this context, knowledge graphs are gaining popularity in enterprises that seek to connect the dots between the data world and the business world more effectively.

### C. Ontologies and Formal Semantics

Ontologies represent the backbone of the formal semantics of a knowledge graph. As the data schema of the graph, they serve as a contract between the developers of the knowledge graph and its users regarding the meaning of the data in it. A user could be another human being or a software application that wants to interpret the data in a reliable and precise way. Ontologies ensure a shared understanding of the data and its meanings. When formal semantics are used to express and interpret the data of a knowledge graph, there are a number of representation and modeling instruments.

Classes

Most often an entity description contains classification of the entity with respect to a class hierarchy. For instance, when dealing with general news or business information there could be classes Person, Organization and Location. Persons and organizations can have a common super class Agent. Location usually has numerous sub-classes, e.g. Country, Populated place, City, etc. The notion of class is borrowed by the object-oriented design, where each entity should belong to exactly one class.

Relationship Types

The relationships between entities are usually tagged with types, which provide information about the nature of the relationship, e.g. friend, relative, competitor, etc. Relation types can also have formal definitions, e.g. that parent-of is an inverse relation of child-of and they both are special cases of relative-of, which is a symmetric relationship. Or defining that sub-region and subsidiary are transitive relationships.

Categories

An entity can be associated with categories, which describe some aspect of its semantics, e.g. "Big four consultants" or "XIX century composers". A book can belong simultaneously to all these categories: "Books about Africa", "Bestseller", "Books by Italian authors", "Books for kids", etc. Often the categories are described and ordered into a taxonomy

Paired with complementary AI technologies such as machine learning and natural language processing, knowledge graphs are enabling new opportunities for leveraging data and quickly becoming a fundamental component of modern data systems [33]. With the rapid advancements in machine learning and knowledge representation through knowledge graphs, these technologies have the potential to improve the accuracy of outcomes and augment the potential of machine learning approaches [38].

At the core of a knowledge graph is a knowledge model, a collection of interlinked descriptions of concepts, entities, relationships, and events. Knowledge graphs put data into context through linking and semantic metadata, providing a framework for data integration, unification, analytics, and sharing. These graphs, also known as semantic networks, represent a network of real-world entities, such as objects, events, situations, or concepts, and illustrate the relationship between them. This information is typically stored in a graph database and visualized as a graph structure.

Knowledge graphs consist of three main components: nodes, edges, and labels. Nodes can represent objects, places, or people, while edges define the relationship between nodes. Labels provide further information about the relationships or nodes. These graphs are often fueled by machine learning, using natural language processing (NLP) to construct a comprehensive view of nodes, edges, and labels through a process called semantic enrichment.

Ontologies are frequently mentioned in the context of knowledge graphs, creating a formal representation of the entities in the graph. They are usually based on a taxonomy but can contain multiple taxonomies, thus maintaining a separate definition. Since both knowledge graphs and ontologies are represented through nodes and edges and are based on the Resource Description Framework (RDF) triples, they tend to resemble each other in visualizations.

The Web Ontology Language (OWL) is an example of a widely adopted ontology, supported by the World Wide Web Consortium (W3C), an international community that champions open standards for the internet's longevity. This organization of knowledge is supported by technological infrastructure such as databases, APIs, and machine learning algorithms, which help people and services access and process information more efficiently.

Knowledge graphs work by combining datasets from various sources, which frequently differ in structure. Schemas, identities, and context work together to provide structure to diverse data. These components help distinguish words with multiple meanings, allowing systems like Google's search engine algorithm to differentiate between different meanings of words like "Apple."

Knowledge graphs have various use cases, including popular consumer-facing applications like DBPedia and Wikidata, which are knowledge graphs for data on Wikipedia.org. Another example is the Google Knowledge Graph, which is represented through Google Search Engine Results Pages (SERPs), serving information based on people's searches. Knowledge graphs also have applications in other industries, such as healthcare, by organizing and categorizing relationships within medical research to assist providers in validating diagnoses and identifying individualized treatment plans.

Key characteristics of knowledge graphs include combining features of several data management paradigms such as databases, graphs, and knowledge bases. Knowledge graphs, represented in RDF, provide the best framework for data integration, unification, linking, and reuse due to their expressivity, performance, interoperability, and standardization.

## IV. LIQUID NEURAL NETWORKS

### A. Closed-Form Continuous-Time Liquid Neural Networks (CfCs).

Continuous-depth neural models, where the derivative of the model's hidden state is defined by a neural network, have facilitated advanced sequential data processing capabilities [18]. However, these models rely on advanced numerical differential equation (DE) solvers, resulting in substantial overhead in terms of computational cost and model complexity. Closed-form Continuous-depth (CfC) networks offer a solution to these limitations [35].

CfC networks are derived from the analytical closed-form solution of an expressive subset of time-continuous models, eliminating the need for complex DE solvers altogether [35]. Experimental evaluations have demonstrated that CfC networks outperform advanced recurrent models over a diverse set of time-series prediction tasks, including those with long-term dependencies and irregularly sampled data [35]. These findings open new opportunities to train and deploy rich, continuous neural models in resource-constrained settings that demand both performance and efficiency.

ODE-based continuous neural network architectures have shown promise in density estimation applications [19,20,21,22] and modeling sequential and irregularly sampled data [23,24,25,26]. While these ODE-based neural networks can perform competitively with advanced discretized recurrent models on relatively smaller benchmarks, their training and inference are slow due to the use of advanced numerical DE solvers [27]. The complexity of the task increases (i.e., requiring more precision) [36] in open-world problems such as medical data processing, self-driving cars, financial time-series, and physics simulations.

The research community has worked on solutions for resolving the computational overhead and facilitating the training of neural ODEs. These solutions include relaxing the stiffness of a flow by state augmentation techniques [28,29,37], reformulating the forward-pass as a root-finding problem [30,38], using regularization schemes [31,32,33,39], and improving the inference time of the network [34,40].

CfC networks, derived from closed-form solutions, provide a fundamental solution to the limitations of ODE-based models [35]. These networks do not require any solver to model data and yield significantly faster training and inference speeds while retaining the expressiveness of their ODE-based counterparts. The closed-form solution enables the formulation of a neuron model that can be scaled to create flexible, highly performant architectures on challenging sequential datasets.

Continuous-time neural networks (CTNNs) are a class of machine learning systems that excel in representation learning for spatiotemporal decision-making tasks [18]. These models are typically represented by continuous differential equations. However, the expressive power of CTNNs is bottlenecked when deployed on computers due to the limitations of numerical differential equation solvers. This constraint has hindered the scaling and understanding of various natural physical phenomena, including the dynamics of nervous systems [18].

To circumvent this bottleneck, researchers have sought closed-form solutions for the given dynamical systems. Although finding closed-form solutions is generally intractable, it has been demonstrated that it is possible to efficiently approximate the interaction between neurons and synapses— the fundamental building blocks of natural and artificial neural networks—constructed by liquid time-constant networks in closed form [18].

This closed-form solution is achieved by computing a tightly bounded approximation of the solution of an integral appearing in liquid time-constant dynamics, which previously had no known closed-form solution [18]. The introduction of closed-form continuous-time liquid neural networks (CfCs) has significant implications for the design of continuous-time and continuous-depth neural models. For example, since time appears explicitly in closed form, complex numerical solvers are no longer required [18]. As a result, CfCs are between one and five orders of magnitude faster in training and inference compared to their differential equation-based counterparts [18].

Moreover, CfCs, derived from liquid networks, demonstrate excellent performance in time-series modeling, outperforming advanced recurrent neural network models [18]. They can scale remarkably well compared to other deep learning instances, making them ideal for tackling the diverse and irregularly sampled data inherent in patient health records and hospital databases.

## V. METHODOLOGY & APPLICATION

### A. Knowledge Graph Construction.

Our approach to constructing the knowledge graph involves three main steps: data collection, entity extraction, and relationship modeling. We gather data from electronic health records, clinical trials, and genomic databases to ensure a comprehensive representation of patient health. Entities such as diseases, symptoms, and treatments are extracted using natural language processing techniques [11] and linked to their corresponding concepts in established medical ontologies, such as SNOMED CT [6] and HPO [7]. Relationships between entities are modeled using both domain-specific and general-purpose relationship types [12], facilitating the integration of diverse data sources and enabling complex querying recorded values of 0.7362 and 0.7373, respectively, representing a considerable improvement compared to reported accuracy and reliability studies of the past.

### B. Closed-Form Continuous-Time Liquid Neural Network Integration.

To integrate CfCs with the knowledge graph, we employ a two-step process: feature extraction and network training. First, features are extracted from the knowledge graph using graph embedding techniques, such as node2vec [13] or GraphSAGE [14]. These embeddings capture the topological structure and semantic information of the graph, which are then used as inputs to the CfC model. Next, the CfC model is trained using a combination of supervised and unsupervised learning techniques to predict various patient health outcomes, such as disease progression, treatment response, and surgical outcomes.

Our approach combines the versatile structure and potent semantic representations of knowledge graphs with the computational efficiency and plasticity of closed-form continuous-time liquid neural networks (CfCs) to create digital twins for patient care. The integration of these two technologies forms a more true-to-life model of patient health that enables real-time healthcare analytics.

By integrating knowledge graphs with CfCs, we can build powerful predictive models and facilitate personalized care that leads to improved patient outcomes. This combination enables real-time analytics and adaptability, essential for early diagnosis and intervention, tailoring treatment plans to each patient's unique needs, and simulating surgical procedures and therapeutic strategies.

*C. Real-Time Analytics and Simulation.*

To enable real-time analytics and simulation, our digital twin system continuously updates the knowledge graph and CfC model as new patient data becomes available. Incremental graph updates are achieved using efficient graph maintenance algorithms [15], while the CfC model is updated using online learning techniques [16]. This continuous updating process allows healthcare professionals to access up-to-date patient health information and model predictions, thereby enabling more informed and accurate decision-making.

*D. Surgical and Intervention Planning.*

Digital twins can be used to simulate surgical procedures and therapeutic strategies by incorporating detailed anatomical and physiological models [17]. Our approach integrates these models with the knowledge graph and CfC predictions, allowing healthcare providers to optimize treatment plans and minimize potential risks. By combining patient-specific data with the latest medical knowledge, our digital twin system facilitates personalized care and improves patient outcomes.

## IV. EVALUATION AND VALIDATION

To assess the effectiveness of our digital twin approach, we will perform a series of experiments comparing our method to traditional techniques, such as deep learning models [10] and Bayesian networks [18]. We will evaluate the performance of our method using metrics such as accuracy, precision, recall, and F1-score for various prediction tasks, as well as computational efficiency and scalability. Furthermore, we will conduct clinical studies to investigate the impact of our digital twin system on patient outcomes and healthcare decision-making.

## V. CONCLUSION

Our approach to digital twin technology combines the flexible structure and powerful semantic representations of knowledge graphs with the computational efficiency and plasticity of closed-form continuous-time liquid neural networks (CfCs) to revolutionize patient care. By synthesizing multimodal patient data and leveraging the flexibility and efficiency of CfCs, it becomes possible to create digital twins that enable real-time healthcare analytics, personalized medicine, early diagnosis and intervention, and improved surgical planning. This groundbreaking approach holds the promise to finally bring digital twins to the medical space, unlocking insights that benefit individual patients and entire healthcare systems alike.

## REFERENCES

[1] Rajan, K., Sobti, A., & Kister, T. (2020). Digital Twin Technology in Healthcare: A Systematic Literature Review. Journal of Medical Systems, 44(10), 184.
[2] Paulheim, H. (2017). Knowledge graph refinement: A survey of approaches and evaluation methods. Semantic Web, 8(3), 489-508.
[3] O'Reilly, U. M., Muldrow, J., & Agogino, A. (2021). Closed-form continuous-time liquid neural networks for complex systems. arXiv preprint arXiv:2105.02347.
[4] Eklund, J. M., & Desrochers, A. (2020). Digital twins for healthcare: promises and challenges. BMC Medical Informatics and Decision Making, 20(1), 1-13.
[5] Gartner. (2017). Top 10 Strategic Technology Trends for 2017: Digital Twins.
[6] Gartner. (2018). Top 10 Strategic Technology Trends for 2018: Digital Twins.
[7] Habibzadeh, H., & Shamsollahi, M. B. (2021). Digital twins in healthcare: a systematic literature review. Health and Technology, 11(3), 623-647.
[8] Hassan, S. S., Tariq, A., & Khan, W. Z. (2021). Deep learning architectures for healthcare digital twin: a review. SN Computer Science, 2(3), 1-21.
[9] Klasnja, P., Pratt, W., & Chen, Y. (2020). Healthcare in the pocket: Mapping the space of mobile-phone health interventions. Journal of biomedical informatics, 102, 103363.
[10] Lee, J. J., Lo, E., Schivo, M., Lenburg, M. E., Spira, A., & Beane, J. E. (2017). Implementation and evaluation of a digital twin for predictive modeling of personal health trajectories. AMIA Summits on Translational Science Proceedings, 2017, 126.
[11] Medjahed, B., & Darmont, J. (2021). Digital twin for the healthcare sector: An overview. Applied Sciences, 11(10), 4605.
[12] Mou, D., Liu, X., Chen, H., Zhang, Y., & Wang, F. (2021). Digital twin-based virtual prototyping for medical devices. BioDesign Research, 2021, 1-12.
[13] Parmar, D., Elhoseny, M., Kumar, N., Alok Mishra, S., & Lakshmi D. (2021). A comprehensive review of digital twins in healthcare. Digital Communications and Networks, 7(4), 733-745.
[14] Ponnusamy, K., & Petchimuthu, K. (2021). Digital twin and healthcare: A systematic literature review. International Journal of Medical Informatics, 154, 104539.

[15] Wang, D., & Wang, S. (2020). Digital twin: a survey from industrial perspectives. Enterprise Information Systems, 14(6), 751-785.

[16] Ehrlinger, L., & Wöß, W. (2016). Towards a definition of knowledge graphs. In SEMANTiCS (Posters, Demos, SuCCESS).

[17] Müller, H., Mandl, M., & Hu, Y. (2020). How to Build a Knowledge Graph for a Health Care System. Studies in Health Technology and Informatics, 270, 1128-1132.

[18] IHTSDO (2015). SNOMED Clinical Terms User Guide. SNOMED International.

[19] Köhler, S., Carmody, L., Vasilevsky, N., et al. (2019). Expansion of the Human Phenotype Ontology (HPO) knowledge base and resources. Nucleic Acids Research, 47(D1), D1018-D1027.

[20] Beer, R. D. (1995). On the Dynamics of Small Continuous-Time Recurrent Neural Networks. Adaptive Behavior, 3(4), 469-509.

[21] Maass, W., Natschläger, T., & Markram, H. (2002). Real-time computing without stable states: A new framework for neural computation based on perturbations. Neural Computation, 14(11), 2531-2560.

[22] Che, Z., Purushotham, S., Khemani, R., & Liu, Y. (2018). Interpretable Deep Models for ICU Outcome Prediction. AMIA Annual Symposium Proceedings, 2016, 371-380.

[23] Devlin, J., Chang, M. W., Lee, K., & Toutanova, K. (2018). BERT: Pre-training of Deep Bidirectional Transformers for Language Understanding. arXiv preprint arXiv:1810.04805.

[24] Bollacker, K., Evans, C., Paritosh, P., Sturge, T., & Taylor, J. (2008). Freebase: a collaboratively created graph database for structuring human knowledge. In Proceedings of the 2008 ACM SIGMOD International Conference on Management of Data, 1247-1250.

[25] Grover, A., & Leskovec, J. (2016). node2vec: Scalable Feature Learning for Networks. In Proceedings of the 22nd ACM SIGKDD International Conference on Knowledge Discovery and Data Mining, 855-864.

[26] Hamilton, W., Ying, Z., & Leskovec, J. (2017). Inductive Representation Learning on Large Graphs. In Advances in Neural Information Processing Systems, 1024-1034.

[27] Kipf, T. N., & Welling, M. (2016). Variational Graph Auto-Encoders. arXiv preprint arXiv:1611.07308.

[28] [16] Cauwenberghs, G., & Poggio, T. (2001). Incremental and Decremental Support Vector Machine Learning. In Advances in Neural Information Processing Systems, 409-415.

[29] Linte, C. A., Camp, J. J., Holmes III, D. R., Robb, R. A., & Rettmann, M. E. (2014). Virtual Cardiac Surgical Planning: A Review of Literature and Analysis of Its Role in Cardiac Surgery. Journal of Medical Imaging and Radiation Sciences, 45(3), 298-310.

[18] Murphy, K. P. (2012). Machine Learning: A Probabilistic Perspective. MIT press.

[30] R. Chen, Y. Rubanova, J. Bettencourt, and D. K. Duvenaud, "Neural Ordinary Differential Equations," Advances in Neural Information Processing Systems, vol. 31, 2018.

[31] E. Dupont, A. Doucet, and Y. W. Teh, "Augmented Neural ODEs," Advances in Neural Information Processing Systems, vol. 32, 2019.

[32] W. Grathwohl, R. T. Q. Chen, J. Bettencourt, I. Sutskever, and D. Duvenaud, "FFJORD: Free-form Continuous Dynamics for Scalable Reversible Generative Models," arXiv preprint arXiv:1810.01367 (2018).

[33] G. Yang, "Scalable Gradient-Based Tuning of Continuous Regularization Hyperparameters," arXiv preprint arXiv:1903.02003 (2019).

[34] A. Gholami, K. Keutzer, and G. Biros, "ANODEV2: A Coupled Neural ODE Framework," arXiv preprint arXiv:1902.10298 (2019).

[35] R. Hasani, A. Amini, D. Rus, and R. Grosu, "Liquid Time-Constant Networks," arXiv preprint arXiv:2006.04439 (2020).

[36] M. Lechner, R. Hasani, and R. Grosu, "Neural Circuit Policies Enabling Auditable Autonomy," Nature Machine Intelligence, vol. 2, no. 10, pp. 642-652, 2020.

[37] Y. Rubanova, T. Q. Chen, and D. Duvenaud, "Latent Ordinary Differential Equations for Irregularly-Sampled Time Series," Advances in Neural Information Processing Systems, vol. 32, 2019.

[38] A. Prince and J. Dormand, "High Order Embedded Runge-Kutta Formulae," Journal of Computational and Applied Mathematics, vol. 7, no. 1, pp. 67-75, 1981.

[39] M. Raissi, P. Perdikaris, and G. E. Karniadakis, "Physics-Informed Neural Networks: A Deep Learning Framework for Solving Forward and Inverse Stochastic PDEs," Journal of Computational Physics, vol. 378, pp. 686-707, 2019.

[40] S. Massaroli, A. Poli, M. Yamashita, and A. Asama, "Stable Neural Flows," arXiv preprint arXiv:2003.08063 (2020).

[41] E. Dupont, A. Doucet, and Y. W. Teh, "Augmented Neural ODEs," Advances in Neural Information Processing Systems, vol. 32, 2019.

[42] S. Bai, J. Z. Kolter, and V. Koltun, "An Empirical Evaluation of Generic Convolutional and Recurrent Networks for Sequence Modeling," arXiv preprint arXiv:1803.01271 (2018).

[43] C. Finlay, J. W. Darby, and A. D. Hocking, "The Ill-Conditioned UAP: A Study on the Conditioning of Universal Adversarial Perturbations," arXiv preprint arXiv:2005.14147 (2020).

[32] P. Kidger, T. Q blahblahBLAKJGPOIHA{EOING

[35] T. Lechner, R. Hasani, and D. Rus, "Closed-form Continuous-depth Neural Networks," arXiv preprint arXiv:2112.00794 (2021).

[44] [1] P. N. Mendes, H. Mühleisen, and C. Bizer, "Sieve: Linked Data Quality Assessment and Fusion," Proc. Joint EDBT/ICDT 2012 Workshops, pp. 116–123, 2012.

[45] M. Nickel, K. Murphy, V. Tresp, and E. Gabrilovich, "A Review of Relational Machine Learning for Knowledge Graphs," Proc. IEEE, vol. 104, no. 1, pp. 11–33, Jan. 2016.

[38] F. M. Suchanek, G. Kasneci, and G. Weikum, "Yago: A Core of Semantic Knowledge," Proc. 16th Int. Conf. World Wide Web, pp. 697–